\documentclass[11pt]{article}

\usepackage[preprint]{acl}

\usepackage{times}
\usepackage{latexsym}

\usepackage{graphicx}
\usepackage[T1]{fontenc}

\usepackage[utf8]{inputenc}

\usepackage{microtype}

\usepackage{inconsolata}

\title{Are LLMs Effective Backbones for Fine-tuning? An Experimental Investigation of Supervised LLMs on Chinese Short Text Matching}

\author{Shulin Liu, Chengcheng Xu, Hao Liu, Tinghao Yu, Tao Yang \\
Machine Learning Platform Department\\
Tencent TEG. Beijing, China \\
\{forestliu, doublecxu, paulhliu, maxwellyu, rigorosyang\}@tencent.com
}

\begin{document}
\maketitle

\begin{abstract}
The recent success of Large Language Models (LLMs) has garnered significant attention in both academia and industry. Prior research on LLMs has primarily focused on enhancing or leveraging their generalization capabilities in zero- and few-shot settings. However, there has been limited investigation into effectively fine-tuning LLMs for a specific natural language understanding task in supervised settings. In this study, we conduct an experimental analysis by fine-tuning LLMs for the task of Chinese short text matching. We explore various factors that influence performance when fine-tuning LLMs, including task modeling methods, prompt formats, and output formats.

\end{abstract}

\section{Introduction}
\label{sec:intro}
The recent success of Large Language Models (LLMs), such as GPT-3\cite{brown2020language}, LLaMA\cite{touvron2023llama} and PaLM\cite{chowdhery2023palm}, has garnered significant attention in both academia and industry. LLMs have demonstrated remarkable generalization capabilities in zero- and few-shot settings, particularly in natural language generation (NLG) tasks. Substantial efforts have been made to enhance and utilizing such generalization capabilities\cite{xu-etal-2023-fine,saad-falcon-etal-2023-udapdr,yun-etal-2023-appraising}.

However, for natural language understanding (NLU) tasks, zero- and few-shot LLMs struggle to achieve satisfactory performance\cite{nie2022improving,wei2023zero,li2023evaluating,li2023label} compared to fine-tuned small models (e.g., Bert base\cite{devlin2018bert}). Our experimental results on the task of Chinese short text matching also confirm this phenomenon. As presented in Section\ref{sec:model}, fine-tuned Bert achieves an accuracy of 84.5\% on the BQ\cite{chen2018bq} corpus, while GPT-4\footnote{The metrics are measured by utilizing OpenAI API.}, one of the most successful LLMs, only attains an accuracy score of 52.9\% in zero-shot and 77.9\% in few-shot settings. There has been limited investigation into effectively tuning LLMs for a specific NLU task in supervised settings. In this paper, we explore various factors affecting the performance of LLMs for Chinese short text matching task, including task modeling methods, prompt formats, and output formats.

\begin{itemize}
\item \textbf{Task modeling methods:} In this study, we examine the impacts of modeling this task as both a generative task and a discriminative classification task, respectively. \textit{(1) Generative Task:} LLMs uniformly model all tasks as generative tasks. Following this principle, we organize the given pair of sentences into a single text as input and make the model generate the target label (equivalent or inequivalent). \textit{(2) Discriminative Classification Task:} Motivated by the efficacy of fine-tuning Bert for text matching\cite{chen2020neural,qi2022all}, we concatenate the given pair of texts as input, extract vector representations from the final LLM layer as features, and perform binary classifications based on the extracted features.
\item \textbf{Prompt Formats:} Prompt design is crucial for LLMs in zero- and few-shot settings\cite{gu2021ppt,liu2023pre}. However, the importance of prompts in supervised settings has not been explored. In this paper, we compare two completely different styles of prompts. One is concise, directly concatenating the given pair of sentences without any explanation of the target task. The other organizes the prompt through complex instructions, including not only the given sentences but also a detail description of the target task.
\item \textbf{Output Formats:} Incorporating the Chain of Thought (CoT) into prompts has been shown to significantly enhance performance in reasoning and complex tasks in zero- and few-shot settings\cite{wei2022chain,wang2022self}. Nevertheless, the impact of CoT on matching tasks in supervised settings has yet to be examined. In this study, we address this gap by \textit{incorporating CoT into the \textbf{output} part} of training samples.
\end{itemize}
We conduct experiments on two widely-used Chinese short text matching datasets, LCQMC \cite{liu2018lcqmc} and BQ \cite{chen2018bq}. All experiments are carried out based on CLLM-7B, which is a Chinese-enhanced model based on LLaMA-2-7B. Our preliminary results demonstrate that the fine-tuned CLLM-7B outperforms both fine-tuned BERT and few-shot GPT-4. Furthermore, the results indicate that the generative paradigm surpasses the discriminative approach, especially when training data is limited. Lastly, our experiments reveal that CoT is also beneficial for the matching task in supervised settings.

\section{Backgrounds}
In this section, we provide a brief overview of the Chinese short text matching task and the datasets employed in this study.

\subsection{Task Definition}
Chinese short text matching, often regarded as a task of identifying sentence semantic equivalence, is a fundamental task of natural language processing. Given a pair of sentences, the goal of a matching model is to ascertain their semantic equivalence. Short text matching is extensively utilized in a range of NLP tasks, such as question answering \cite{liu2018improved} and dialogue systems \cite{pang2008foundations}.
\subsection{Datasets and Metrics}
We conduct experiments on two widely-used Chinese short text matching corpora: LCQMC \cite{liu2018lcqmc} and BQ \cite{chen2018bq}.

LCQMC is a large-scale, open-domain question matching corpus. It comprises 260,068 Chinese search query pairs, including 238,766 training samples, 8,802 development samples, and 12,500 test samples. Each pair is annotated with a binary label indicating whether the two queries share the same intention.

BQ is a domain-specific, large-scale corpus for bank question matching. It consists of 120,000 Chinese sentence pairs, including 100,000 training samples, 10,000 development samples, and 10,000 test samples. Each pair is also annotated with a binary label indicating whether the two sentences convey the same meaning.

We employ accuracy (ACC.) as the evaluation metric, which is the percentage of correctly predicted examples.

\begin{figure*}[t]
\center
  \includegraphics[width=0.9\textwidth]{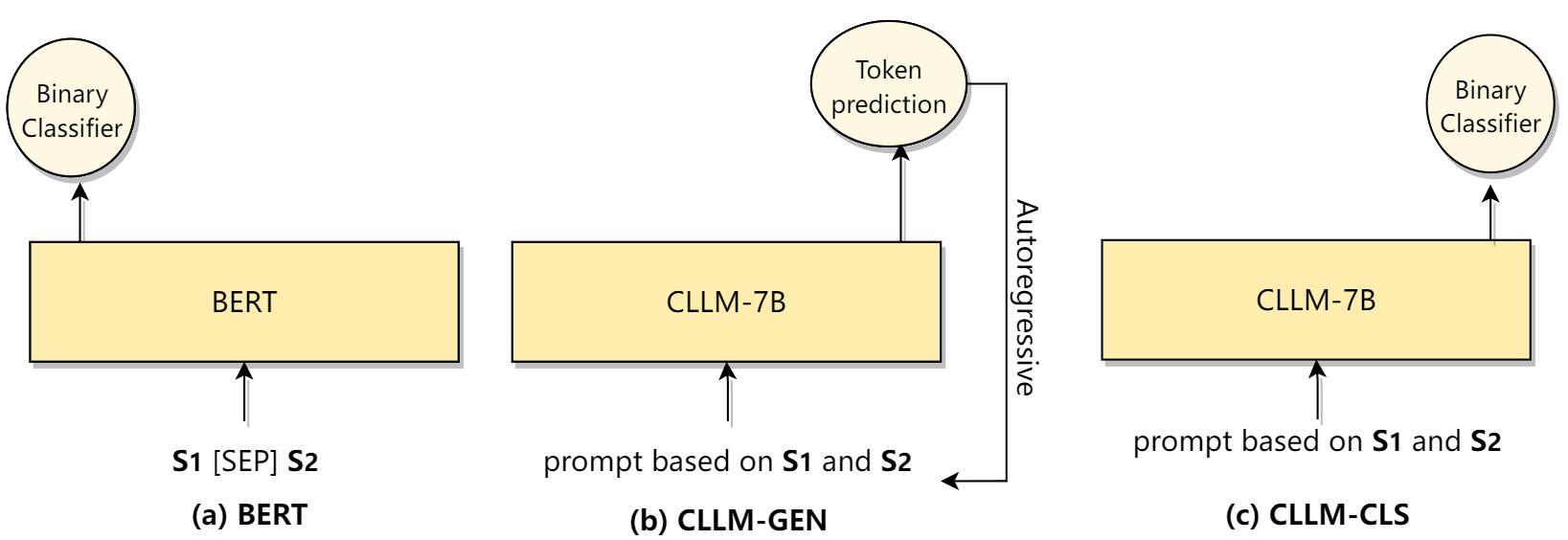}
  \caption{\label{Figure:Modeling} Model structures of modeling text matching as generative and discriminant task.}
\end{figure*}

\section{Experiments and Results}
In this section, we outline the experimental configurations and present the results. We examine the influence of the three factors discussed in Section \ref{sec:intro} through the following experiments. We tune models via full-model fine-tuning.

\subsection{Generative vs. Discriminative Models}
\label{sec:model}
We first outline our approach to fine-tuning LLMs by modeling the matching task as both a generative task and a discriminative task. Subsequently, we present the results and provide an analysis.

\textbf{Modeling as A Generative Task:} LLMs consistently treat all tasks as generative tasks. In line with this principle, we merge the provided pair of sentences with instructions into a single text input and prompt the model to generate the target label. We refer to this model as CLLM-7B-GEN. Figure \ref{Figure:Modeling}(b) illustrates the model structure. We optimize it by maximizing the generation probability of the target label.

\textbf{Modeling as A Discriminative Task:} Inspired by the effectiveness of fine-tuning BERT for text matching tasks (see Figure \ref{Figure:Modeling}(a)), we concatenate the given pair of texts as input, extract vector representations from the final LLM layer as features, and perform binary classification based on the extracted features. We refer to this model as CLLM-7B-CLS. Figure \ref{Figure:Modeling}(c) demonstrates the model structure.

\begin{figure}[t]
\center
  \includegraphics[width=0.49\textwidth]{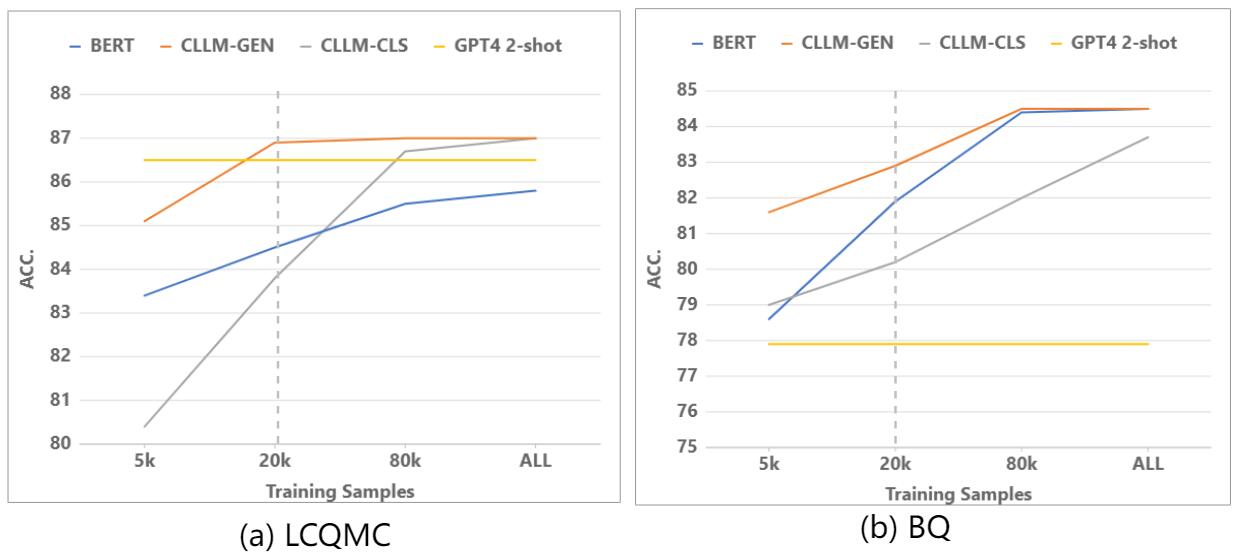}
  \caption{\label{Figure:exp1}  The results of models trained on 5,000, 20,000, 80,000 samples as well as trained on the entire training set.}
\end{figure}

We validated the performance of generative and discriminative models on training sets of different scales. Figure \ref{Figure:exp1} shows the experimental results, where the 2-shot GPT-4 results are measured by calling the official OpenAI API. Figure \ref{Figure:lcqmc-2-shot} and Figure \ref{Figure:bq-2-shot} in Appendix \ref{sec:appendix} illustrate the 2-shot prompts for LCQMC and BQ, respectively. From the results, we observe that:

1) When the number of training samples is less than 20,000, CLLM-GEN significantly outperforms discriminative models, including BERT and CLLM-CLS, on both LCQMC and BQ. This phenomenon is quite intuitive, as the generative approach aligns with the pre-training procedure, making it easier to activate the knowledge acquired by the model during pre-training. Furthermore, due to the massive amount of data used in the pre-training phase of LLMs, the issue of evaluation data leakage cannot be ignored \cite{yang2023rethinking,zhou2023don}. To determine whether CLLM-7B has a data leakage problem, we conducted zero-shot experiments on it. The model achieves an accuracy of 52.1\% on LCQMC and 52.9\% on BQ, slightly better than the 50\% expected from random guessing. Consequently, we believe that both BQ and LCQMC are not included in CLLM-7B's pre-training data.

2) The performance of 2-shot GPT-4 on BQ is much worse than that of supervised models. This is mainly because BQ is a dataset of real customer service questions from WeBank Inc., and a full understanding of the sentences' meaning requires background information about this bank. For example, questions in BQ usually mention specific products or a particular function in the bank's app. This background knowledge is unknown to CLLM and is also impossible to provide entirely in the prompt.

3) CLLM-GEN trained on the whole training corpus on LCQMC outperforms BERT. However, it fails on the BQ corpus. We believe the reason is that CLLM-7B, like BERT, also lack knowledge of WeBank, and such knowledge can only be obtained from the training data. Therefore, compared to BERT, CLLM-7B does not have an advantage on this dataset.

The above experiments demonstrate that generative paradigm is better for supervised LLMs. Therefore, all subsequent experiments will be conducted following this paradigm.

\subsection{Concise vs. Complex Prompts}
\label{sec:complex}
\begin{figure}[t]
\center
  \includegraphics[width=0.49\textwidth]{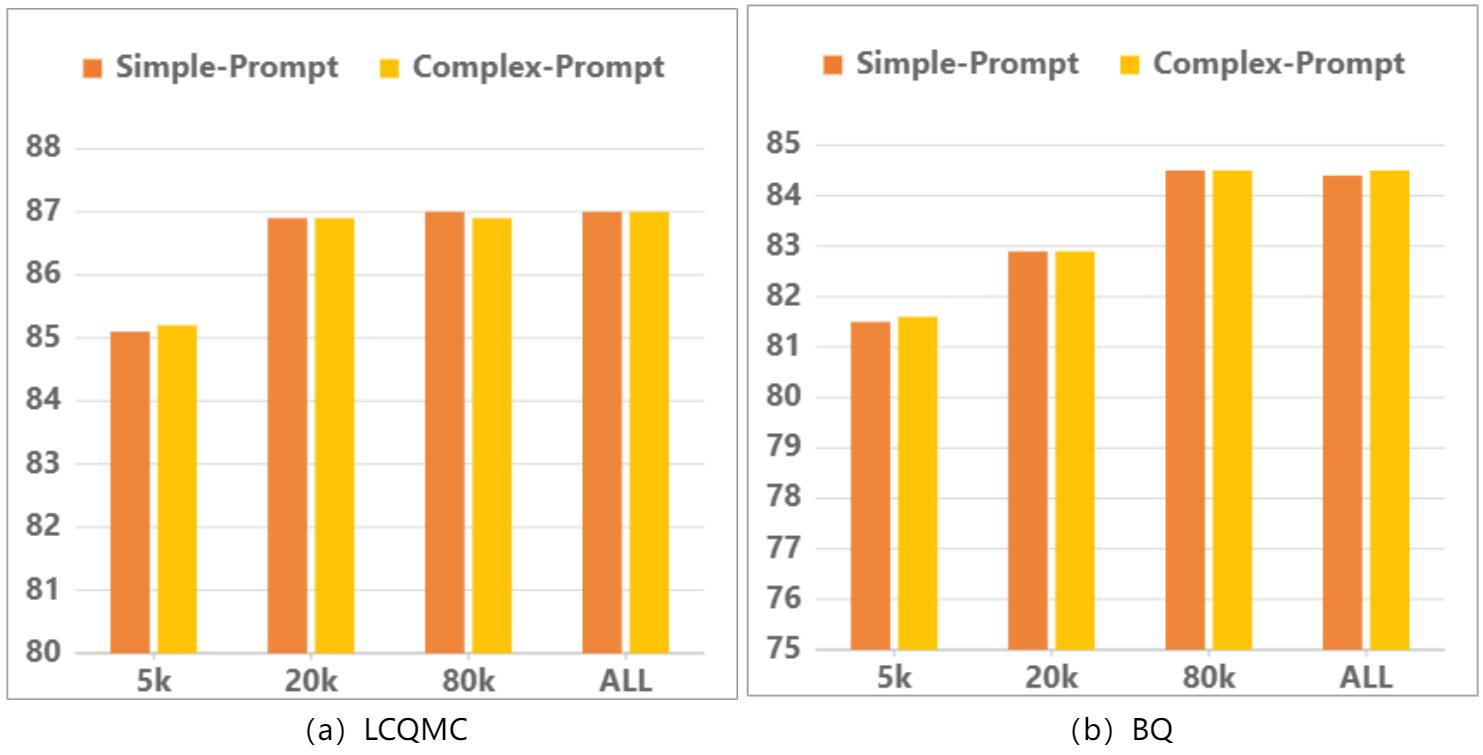}
  \caption{\label{Figure:prompt} The results of concise and complex prompts. }
\end{figure}

Prompt design is crucial for LLMs in zero- and few-shot settings. However, the significance of prompts in supervised settings remains unexplored. In this subsection, we compare two distinct styles of prompts. The concise prompt involves directly concatenating the given text pairs without any explanation of the target task, while the complex prompt organizes the prompt with detailed instructions, incorporating not only the given texts but also a specific description of the target task. Examples of these prompts can be found in Figure \ref{Figure:complex-simple} in Appendix \ref{sec:appendix}.

Figure \ref{Figure:prompt} presents the results, showing that models separately trained by concise and complex prompts achieve comparable performance. This observation suggests that supervised LLMs are not sensitive to prompts. The primary function of a complex prompt is to enhance the model's comprehension of the target task. In supervised scenarios, the model can learn the task definition more accurately from the training data, rendering the prompt design less impactful.

\subsection{Effects of CoT}

\begin{figure}[t]
\center
  \includegraphics[width=0.47\textwidth]{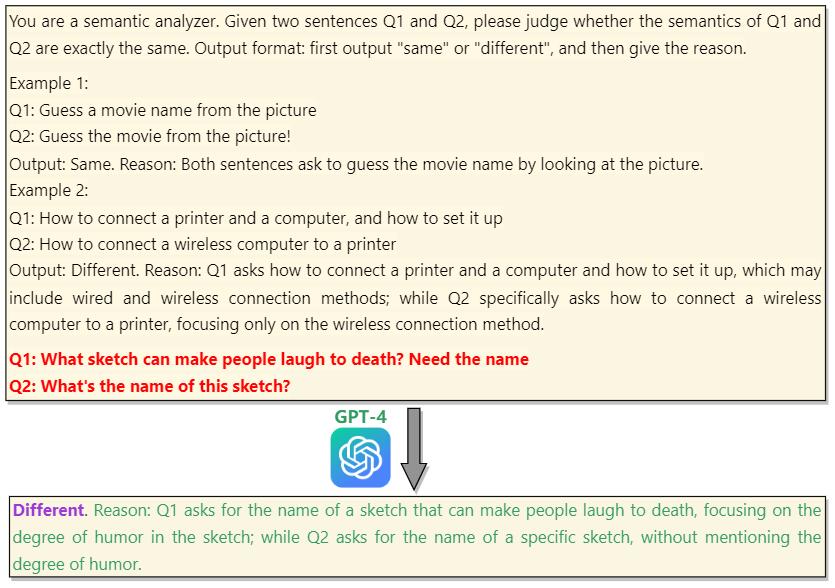}
  \caption{\label{Figure:cot}  Illustration of how to obtain CoT via GPT-4. \textit{All original texts in this figure are in Chinese. For ease of reading, we translated them. The original version is illustrated in Figure \ref{Figure:cot-chinese} in Appendix.}}
\end{figure}

CoT has demonstrated its effectiveness in reasoning and complex tasks within zero- and few-shot settings. However, its efficacy for language understanding tasks in supervised settings remains unexplored.

We have already demonstrated in Section \ref{sec:complex} that adding additional information to the prompt does not improve performance in the supervised setting. Therefore, unlike in zero/few-shot settings, we did not include CoT in the prompt, but instead added it to the output section. Figure \ref{Figure:cot-train-sample} in Appendix A presents a training sample with CoT.

Matching datasets provide labels without CoT. To obtain CoT for the training set, we enlist GPT-4 to determine whether a given pair of texts is equivalent, while also providing explanations for its decision. For samples where GPT-4's judgment aligns with the golden label, we utilize the explanation as the CoT. Conversely, for inconsistent samples, we retain only golden label. Figure \ref{Figure:cot} depicts the designed prompt and response generated by GPT-4. Note that only the output portion of the training samples requires the addition of CoT. Figure \ref{Figure:cot-train-sample} in Appendix presents a training sample that includes CoT. During the evaluation process, we disregard the CoT generated by the model, focusing solely on the label "same" or "different".

In order to reduce the cost, we did not obtain CoT for the entire training set. Instead, we separately sampled 10,000 instances from each dataset and requested GPT-4 to generate CoT. After filtering samples with inconsistent judgments, approximately 86\% of samples in LCQMC and 78\% in BQ retained CoT.

\begin{figure}[t]
\center
  \includegraphics[width=0.47\textwidth]{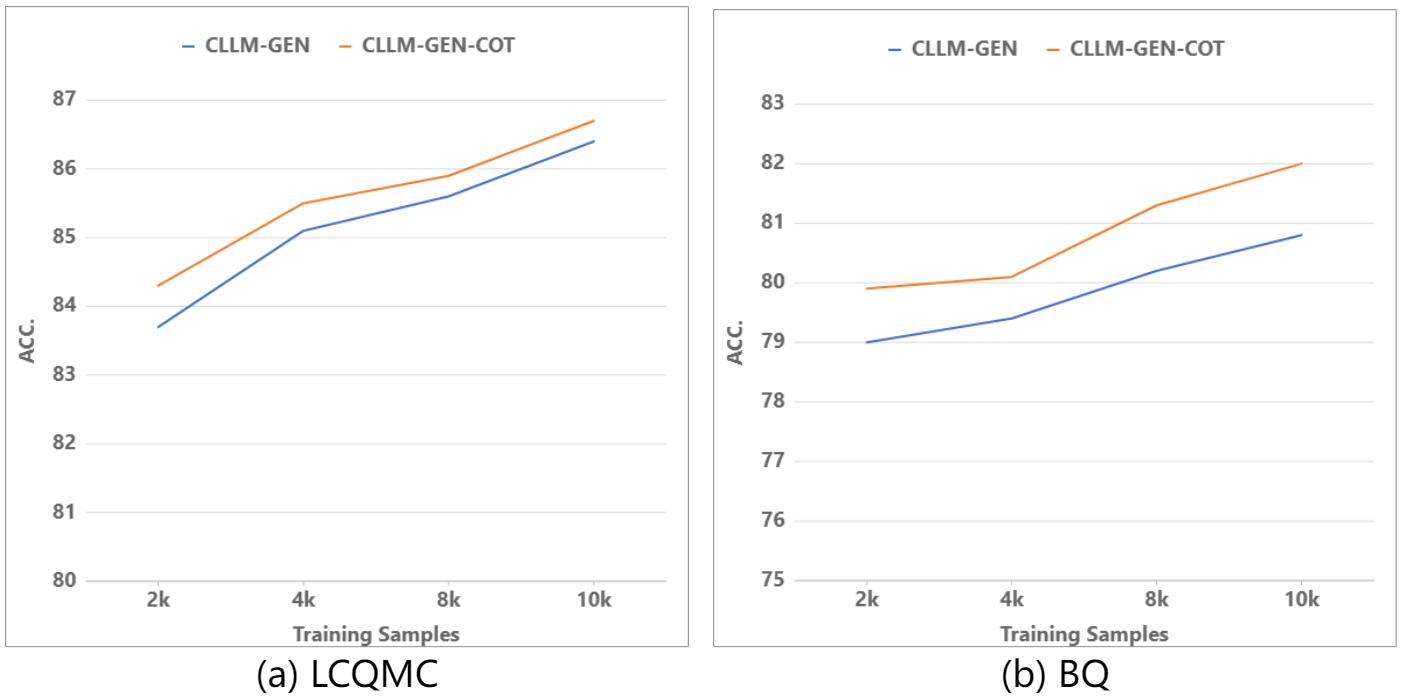}
  \caption{\label{Figure:cot-res} Results of models trained with CoT. }
\end{figure}

We conducted experiments on training sets of varying scales. Figure \ref{Figure:cot-res} displays the results, from which we observe that CoT improves performance on both LCQMC and BQ. Furthermore, the BQ dataset is more challenging than LCQMC, and CLLM-GEN-CoT achieved a more substantial improvement on BQ. This finding suggests that CoT may be particularly effective for difficult tasks.

\section{Conclusions}
In this work, we conduct an experimental study by fine-tuning LLMs on the task of Chinese short text matching. We investigate various factors affecting performance in tuning LLMs, including task modeling methods, prompt formats, and the chain of thought. We systematically carry out experiments on two widely used datasets. The results reveal several insights. First, the fine-tuned CLLM-7B outperforms both fine-tuned BERT and few-shot GPT-4, indicating that LLMs serve as effective backbones in supervised scenarios. Moreover, the generative paradigm is superior to the discriminative one, particularly when training data is limited. Second, supervised LLMs are insensitive to prompts, unlike zero- and few-shot LLMs. Third, CoT is also beneficial for supervised text matching. Although our experiments focus on the task of text matching, the observations may be applicable to other NLU tasks, such as text classification.

\section*{Limitations}
This study has two primary limitations: (1) Prompt engineering is crucial for zero- and few-shot LLMs. We assessed the few-shot performance of GPT-4, as depicted in Figure \ref{Figure:exp1}. Despite our meticulous design of the few-shot prompts, the prompt designs remain subjective and may not necessarily represent the most optimal choices. (2) This study concentrates on the text matching task. Additional experiments might be required to adequately demonstrate if the conclusions drawn in this article are applicable to other NLU tasks (e.g. text classification).

\bibliography{acl_latex}

\appendix

\section{Appendix}
\label{sec:appendix}

\begin{figure}[ht]
\center
  \includegraphics[width=0.45\textwidth]{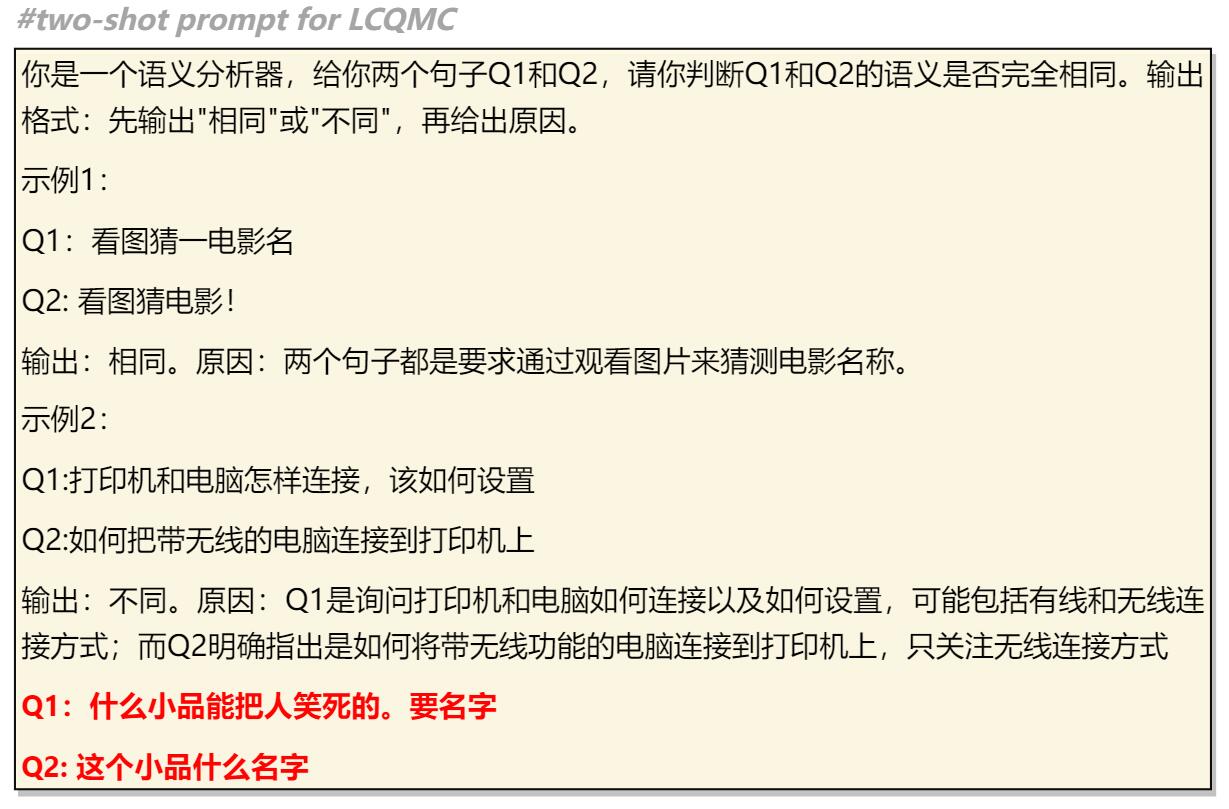}
  \caption{\label{Figure:lcqmc-2-shot} An illustration of 2-shot prompt for LCQMC.}
\end{figure}

\begin{figure}[t]
\center
  \includegraphics[width=0.45\textwidth]{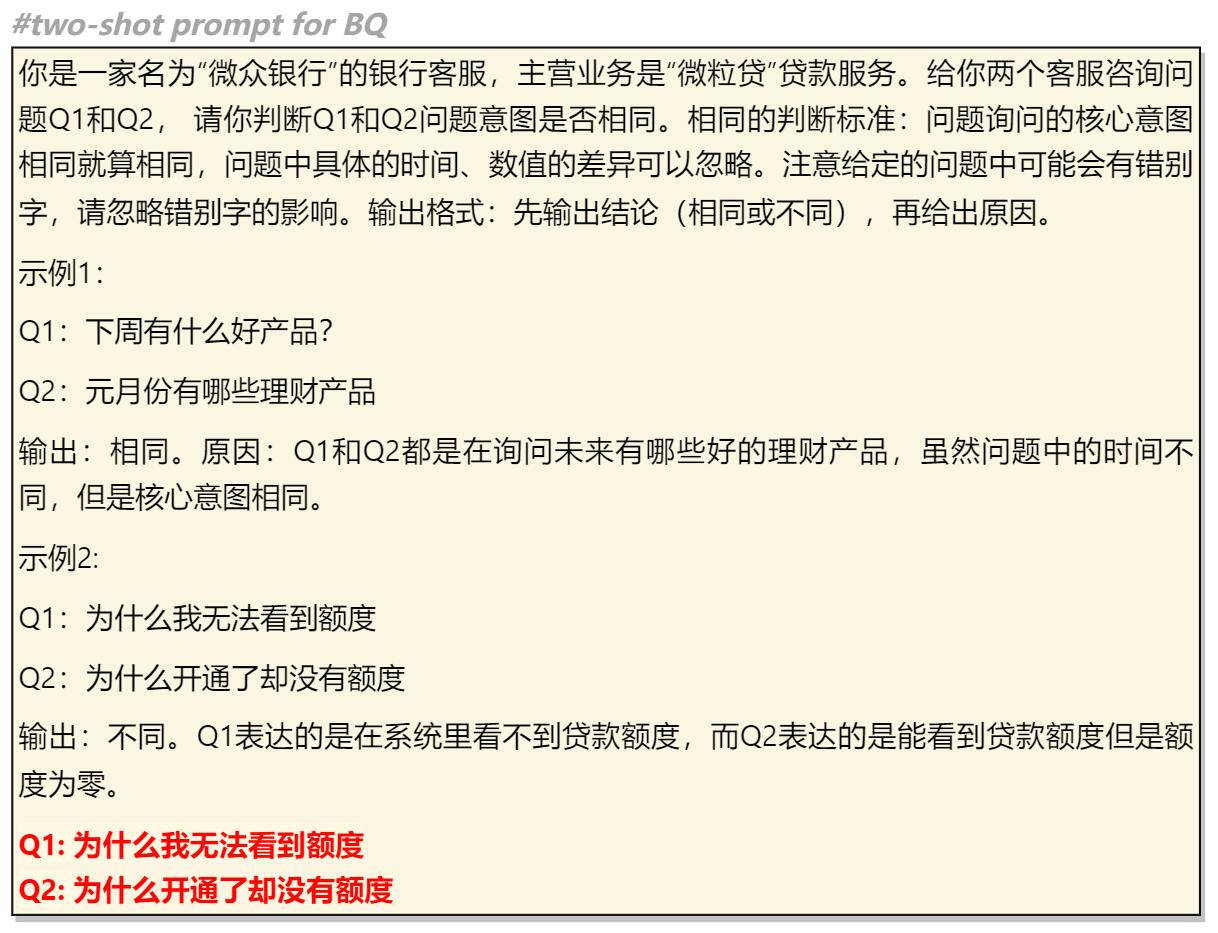}
  \caption{\label{Figure:bq-2-shot} An illustration of 2-shot prompt for BQ.}
\end{figure}

\begin{figure}[t]
\center
  \includegraphics[width=0.45\textwidth]{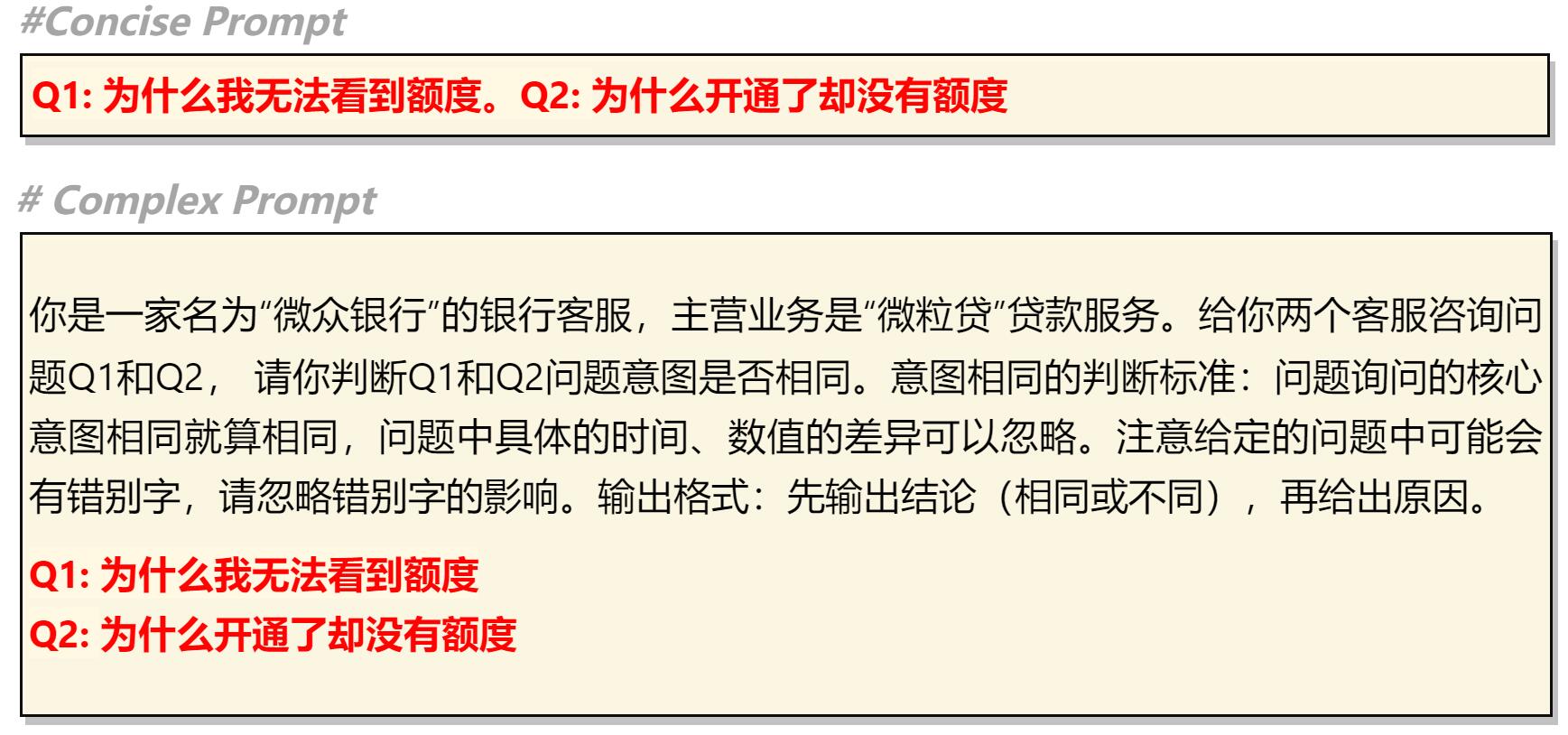}
  \caption{\label{Figure:complex-simple} Examples of complex and simple prompts in Section\ref{sec:complex}}
\end{figure}

\begin{figure}[t]
\center
  \includegraphics[width=0.45\textwidth]{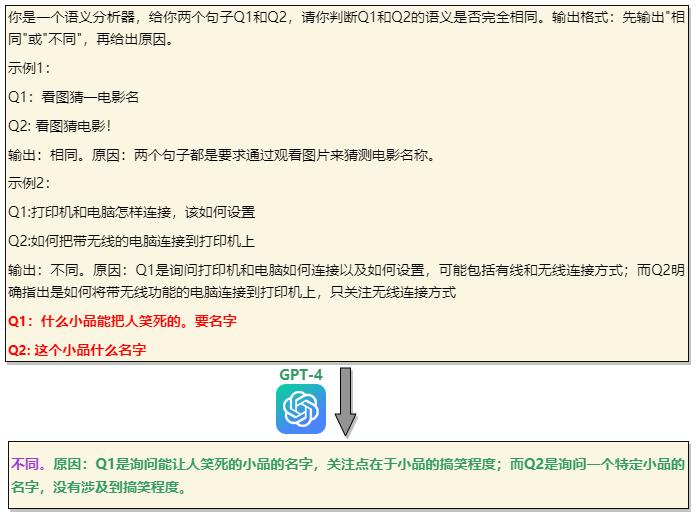}
  \caption{\label{Figure:cot-chinese} The Chinese version of texts in Figure \ref{Figure:cot}}
\end{figure}

\begin{figure}[t]
\center
  \includegraphics[width=0.45\textwidth]{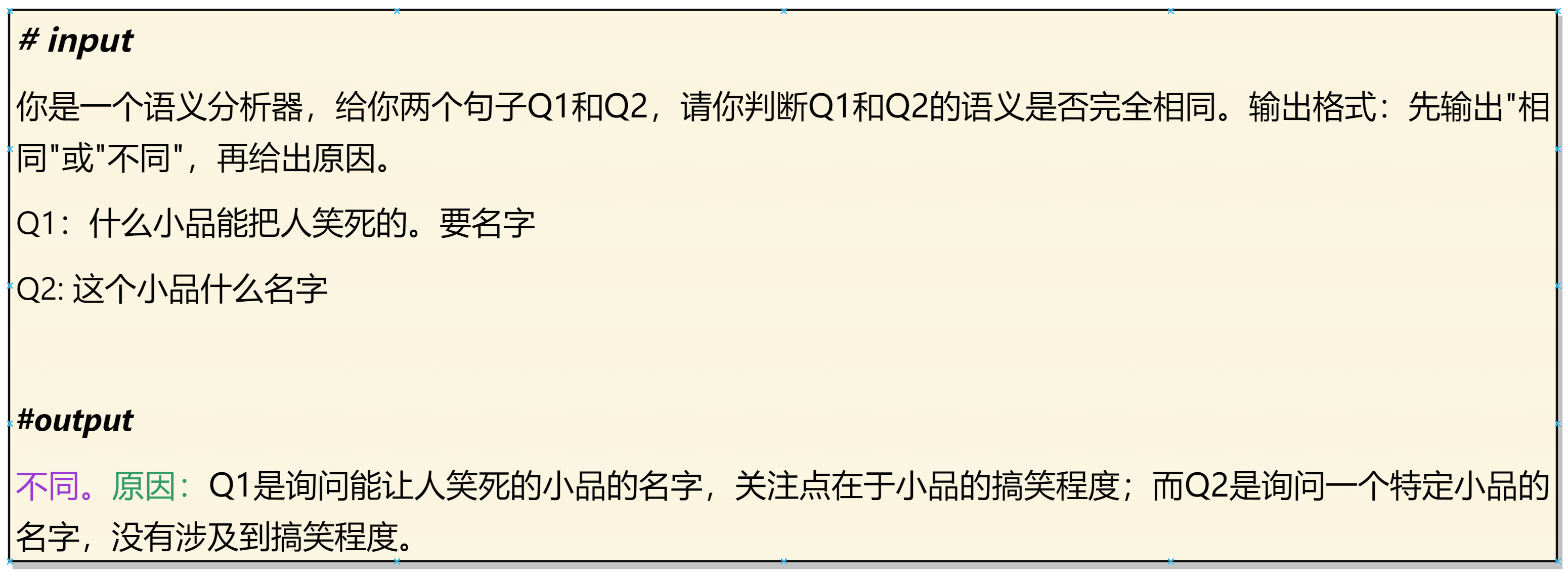}
  \caption{\label{Figure:cot-train-sample} An example of training sample with CoT.}
\end{figure}
\end{document}